\documentclass{article}
\usepackage{spconf,amsmath,graphicx,hyperref,booktabs}

\title{PSMP-CLIP: PATCH-PROMPT SAM AND MULTI-SEMANTIC PROMPTING FOR CLIP-BASED ZERO-SHOT ANOMALY DETECTION}
\name{
Xuezhi Xiang\textsuperscript{1,2,*}, Guanghao Wu\textsuperscript{1}, Heqi Xiang\textsuperscript{3}, Jiayao Liu\textsuperscript{1}, Xiaoheng Li\textsuperscript{1}, Yiming Chen\textsuperscript{1}, Shanjun Zhang\textsuperscript{4}\thanks{This work was supported in part by National Natural Science Foundation of China under Grant 62671193 and 62271160, in part by Heilongjiang Provincial Key R\&D Program Project under Grant 2026ZX01A14, in part by the Natural Science Foundation of Heilongjiang Provincial of China under Grant XQ2026F018, in part by the Fundamental Research Funds for the Central Universities of China under Grant 3072026LJ0802.}
}
\address{
\textsuperscript{1}Information and Communication Engineering, Harbin Engineering University, Harbin, China\\
\textsuperscript{2} 
Key Laboratory of Advanced Marine Communication and Information Technology, Harbin, China \\
\textsuperscript{3}Department of Computer Science, University of Toronto, Toronto, ON M5S 2E4, Canada\\
\textsuperscript{4}The Department of Computer Science, Kanagawa University, Kanagawa, 221-8686, Japan\\
E-mails: xiangxuezhi@hrbeu.edu.cn, wuguanghao@hrbeu.edu.cn, claire.xiang@mail.utoronto.ca
}
\begin{document}
%\ninept
%
\maketitle
\begin{abstract}
Zero-shot anomaly detection aims to localize anomalies without target-domain samples. Existing CLIP-based methods suffer from coarse anomaly maps and limited semantic prompts. We propose PSMP-CLIP, integrating patch-prompt SAM2 segmentation (PPSS) and multi-semantic guided prompt regularization (MSGPR). PPSS samples prompts directly from intermediate patch features, avoiding threshold drift and guiding SAM2 to produce precise masks. MSGPR uses multiple learnable prompts constrained by semantic anchors to preserve generalization. Experiments on 14 datasets show highly competitive performance, achieving the best pixel-level AUROC on MVTec AD, BTAD, DTD-Synthetic, CVC-ClinicDB, TN3K, Endo, and Kvasir.
\end{abstract}
\begin{keywords}
Zero-shot anomaly detection, CLIP, SAM, prompt learning, multi-semantic guided prompt regularization
\end{keywords}
\section{Introduction}
\label{sec:intro}

Anomaly detection is critical in industrial inspection and medical imaging, where anomalous samples are scarce and expensive to annotate. Zero-shot anomaly detection (ZSAD)~\cite{jeong2023winclip} aims to identify and localize anomalies without target-domain training data. Vision-language models such as CLIP~\cite{radford2021learning} provide strong semantic alignment, offering a promising foundation for ZSAD. However, existing CLIP-based methods still face two key limitations: coarse anomaly maps with imprecise boundaries, and prompt representations that converge to a narrow semantic subspace.

CLIP-based zero-shot anomaly detection approaches, including AnomalyCLIP~\cite{zhou2024anomalyclip}, AdaCLIP~\cite{cao2024adaclip}, AA-CLIP~\cite{ma2025aaclip}, Bayes-PFL~\cite{qu2025bayesian}, and MRAD~\cite{xu2026mrad}, align normal/abnormal text prompts with image features to produce anomaly maps. While these methods improve generalization, their outputs remain coarse and boundary-insensitive. To enhance localization, SAM~\cite{kirillov2023segment} and SAM2~\cite{ravi2025sam2} are combined with CLIP. ClipSAM~\cite{li2025clipsam} extracts spatial prompts by applying pixel-level thresholding to CLIP's coarse segmentation maps. However, such strategies rely on thresholding low-resolution outputs, leading to prompt drift when anomalies are absent or subtle. In prompt learning, MSGCoOp~\cite{wang2025msgcoop} adapts CLIP via multi-semantic guided prompts, but it is not specifically adapted for the anomaly detection task, making it hard to adequately meet the demand of anomaly detection.

To address these issues, we propose PSMP-CLIP, a collaborative network that integrates patch-prompt SAM2 segmentation (PPSS) and multi-semantic guided prompt regularization (MSGPR). PPSS directly samples spatial prompts from intermediate patch features, avoiding up-sampling and thresholding on coarse masks. MSGPR uses four learnable prompt groups constrained by semantic anchors, where four groups of anchors are generated for normal and abnormal texts, respectively, to preserve generalization and enrich task-specific semantics.

Our contributions are as follows:
\begin{itemize}
    \setlength{\topsep}{0pt}
    \setlength{\itemsep}{0pt}
    \setlength{\parsep}{0pt}
    \setlength{\parskip}{0pt}
    \item[1)] We propose PPSS, which directly samples spatial prompts from intermediate patch features to avoid prompt coordinate drift, improving fine-grained anomaly localization.
    \item[2)] We propose MSGPR, which uses multiple learnable prompt groups constrained by semantic anchors to suppress catastrophic forgetting and enrich task-specific semantics.
    \item[3)] We conduct comprehensive experiments on 14 datasets. The results show that our method achieves highly competitive performance on pixel-level metrics under the zero-shot setting.
\end{itemize}

% ===== Figure 1: framework (跨栏，放在顶部) =====
\begin{figure*}[t]
    \centering
    \includegraphics[width=\textwidth]{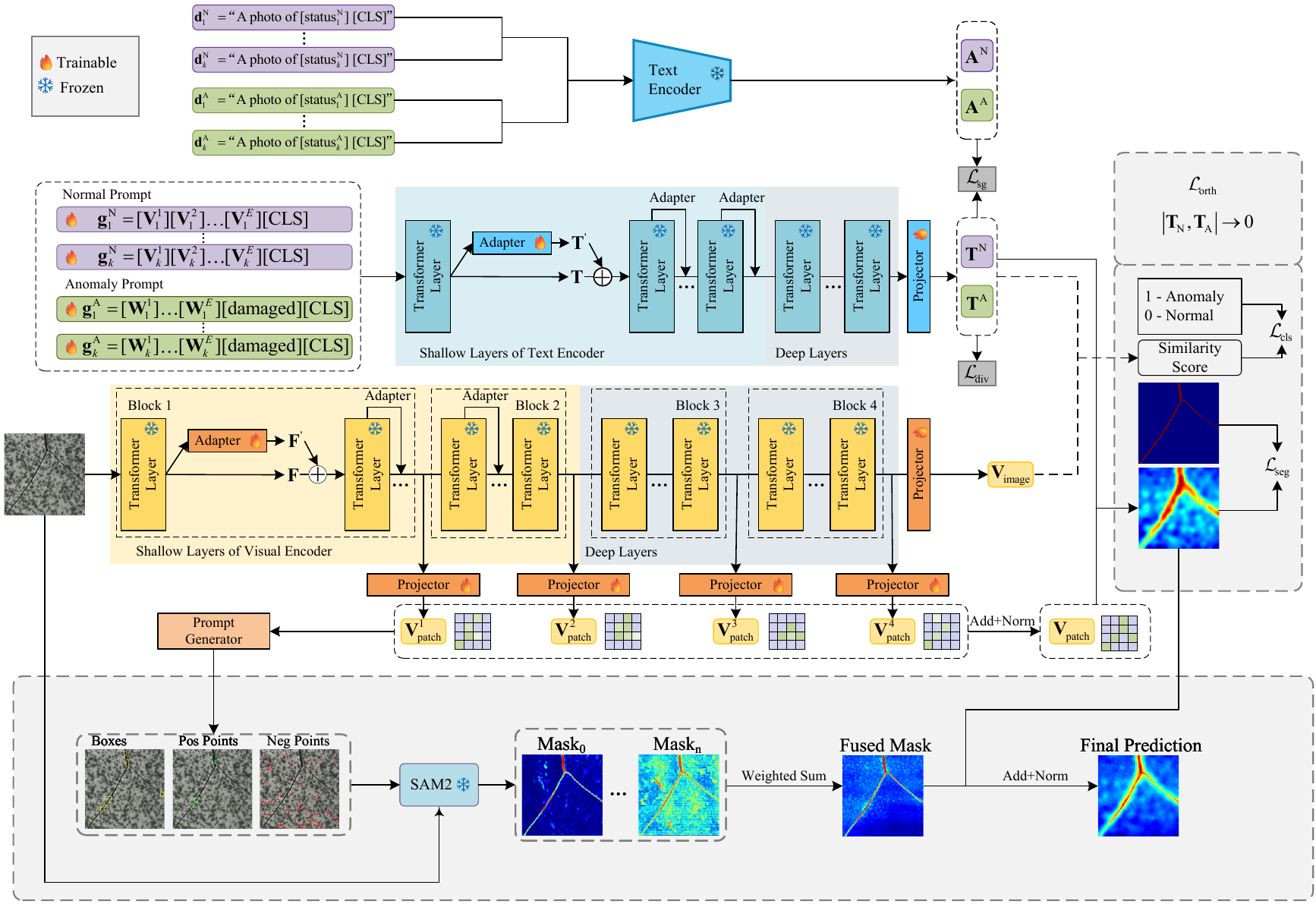}
    \caption{Overview of the proposed PSMP-CLIP network. PPSS enables SAM2 to produce precise boundary segmentations and MSGPR preserves CLIP's general knowledge and enriches task-specific semantics. }
    \label{fig:framework}
\end{figure*}

\section{Method}
\label{sec:method}

\subsection{Overview}
Fig.~\ref{fig:framework} illustrates the overall architecture. PSMP-CLIP builds on AA-CLIP as baseline and comprises two modules: PPSS and MSGPR. During training, the model jointly optimizes classification, segmentation, diversity, semantic guidance, and orthogonality losses. At inference, the final anomaly map is obtained by fusing CLIP coarse prediction with SAM2 fine segmentation.

\subsection{Patch-Prompt SAM2 Segmentation (PPSS)}
\label{ssec:ppss}

\textbf{Multi-level patch feature extraction.} From the 6th, 12th, 18th, and 24th layers of the CLIP image encoder, we extract features $\mathbf{F}_i$ and align them to text dimension via trainable projections, then aggregate:
\begin{equation}
\mathbf{V}_{\mathrm{patch}} = \sum_{i=1}^{4} \mathrm{Proj}_i(\mathbf{F}_i).
\end{equation}

\textbf{Patch-level anomaly scoring.} Compute cosine similarity between $\mathbf{V}_{\mathrm{patch}}$ and the fused normal/abnormal text embeddings $\mathbf{T}^{\mathrm{N}}/\mathbf{T}^{\mathrm{A}}$:
\begin{equation}
\mathbf{p}_{\mathrm{seg}}^{\mathrm{o}} = \cos\left(\mathbf{V}_{\mathrm{patch}} \cdot [\mathbf{T}^{\mathrm{N}}, \mathbf{T}^{\mathrm{A}}]\right),
\end{equation}
and define anomaly score $\mathbf{s}(n)=\mathbf{p}_{\mathrm{seg}}^{\mathrm{A}}(n)-\mathbf{p}_{\mathrm{seg}}^{\mathrm{N}}(n)$.

\textbf{Hybrid prompt generation.} We sample positive points where $\mathbf{s}(n)>\tau$ and negative points where $\mathbf{s}(n)<-\tau$ directly on the patch grid:
\begin{equation}
\mathcal{P}^{\mathrm{patch}} = \bigl\{ \mathbf{p}_n=1 \mid \mathbf{s}(n) > \tau \bigr\} \cup \bigl\{ \mathbf{p}_n=0 \mid \mathbf{s}(n) < -\tau \bigr\}.
\end{equation}

The coordinates of each patch token are mapped to the original pixel space by taking the geometric center:
\begin{equation}
x_n^{\mathrm{c}} = \left(u_n + \frac{1}{2}\right) \frac{W}{W_{\mathrm{patch}}}, \qquad
y_n^{\mathrm{c}} = \left(v_n + \frac{1}{2}\right) \frac{H}{H_{\mathrm{patch}}}.
\end{equation}

We compute connected components of positive points in the patch grid, take the minimum bounding rectangle of each component, and map these rectangles to the original image as box prompts.

\begin{table*}[htbp]
    \centering
    \caption{Pixel-level ZSAD (P-AUROC\% / PRO\%)}
    \label{tab:pixel}
    \small
    \begin{tabular*}{\textwidth}{@{\extracolsep{\fill}}lccccccc@{}}
        \toprule
        Domain & Datasets & AnomalyCLIP & AdaCLIP & AA-CLIP & Bayes-PFL & MRAD & Ours \\
        \midrule
        & MVTec AD & (91.1,81.4) & (86.8,33.8) & (90.6,85.0) & (91.9,\textbf{87.8}) & (\underline{93.0},\underline{86.8}) & (\textbf{93.7},{85.9}) \\
        & VisA & (95.5,87.0) & (95.1,71.3) & (94.5,80.0) & (\underline{95.7},\textbf{90.1}) & (\textbf{95.9},\underline{88.0}) & (95.4,87.0) \\
        Industrial & BTAD & (94.2,\underline{74.8}) & (87.7,17.1) & (95.2,72.3) & (93.9,\textbf{76.6}) & (\underline{95.4},72.8) & (\textbf{95.5},\underline{74.8}) \\
        & MPDD & (95.9,85.5) & (-,-) & (95.7,86.2) & (-,-) & (\textbf{97.9},\textbf{90.6}) & (\underline{96.1},\underline{87.3}) \\
        & DTD-Synthetic & (97.9,\underline{92.3}) & (94.1,24.9) & (96.6,84.1) & (97.8,\textbf{94.3}) & (\underline{98.1},89.8) & (\textbf{98.6},90.0) \\
        & DAGM & (95.6,\underline{91.0}) & (97.0,40.9) & (93.1,82.7) & (\textbf{99.3},\textbf{98.0}) & (\underline{97.4},90.3) & (96.6,89.0) \\
        \midrule
        & CVC-ClinicDB & (82.9,67.8) & (83.6,11.5) & (\underline{88.1},\underline{74.7}) & (86.1,69.8) & (87.3,73.9) & (\textbf{88.2},\textbf{75.2}) \\
        & CVC-ColonDB & (81.9,71.3) & (78.5,10.1) & (82.9,73.5) & (80.6,72.6) & (\textbf{84.7},\underline{73.9}) & (\underline{83.2},\textbf{75.1}) \\
        Medical & TN3K & (81.5,\underline{50.4}) & (82.1,49.8) & (\underline{81.6},47.7) & (-,-) & (-,-) & (\textbf{84.8},\textbf{51.9}) \\
        & Endo & (84.1,63.6) & (84.5,12.1) & (\underline{90.4},\underline{75.7}) & (84.8,63.2) & (88.3,71.6) & (\textbf{91.3},\textbf{77.8}) \\
        & Kvasir & (78.9,45.6) & (-,-) & (\underline{86.7},55.1) & (85.4,\textbf{63.9}) & (84.3,52.7) & (\textbf{89.3},\underline{56.3}) \\
        \bottomrule
    \end{tabular*}
\end{table*}

\textbf{Mask confidence-weighted fusion.} SAM2 produces $K$ candidate masks with confidence scores $\{S_k\}$. After softmax normalization, the fused SAM2 probability map is:
\begin{equation}
\mathbf{P}_{\mathrm{SAM2}} = \sum_{k=1}^{K} \frac{\exp(S_k)}{\sum_{i=1}^{K}\exp(S_i)} \mathbf{P}_k,
\end{equation}
where $\mathbf{P}_k = \mathrm{Sigmoid}(\mathbf{Z}_k)$. The final result is $\mathbf{P}_{\mathrm{final}} = (1-w_0)\mathbf{P}_{\mathrm{CLIP}} + w_0\mathbf{P}_{\mathrm{SAM2}}$, with $w_0=0.8$.

\subsection{Multi-Semantic Guided Prompt Regularization (MSGPR)}
\label{ssec:msgpr}

We construct $k{=}4$ parallel learnable prompt groups for normal and abnormal texts. Each group is encoded by the adapter-equipped text encoder, and the fused features are obtained by averaging:
\begin{equation}
\mathbf{T}^{\mathrm{N}} = \frac{1}{k}\sum_{i=1}^{k}\mathbf{t}_i^{\mathrm{N}}, \qquad
\mathbf{T}^{\mathrm{A}} = \frac{1}{k}\sum_{i=1}^{k}\mathbf{t}_i^{\mathrm{A}}.
\end{equation}

To preserve general knowledge, we build $k$ fixed semantic anchors for normal and abnormal states from templates like ``A photo of [status] [CLS]'' and encode them with the frozen CLIP text encoder. The averaged anchors are:
\begin{equation}
\mathbf{A}^{\mathrm{N}} = \frac{1}{k}\sum_{i=1}^{k}\mathbf{a}_i^{\mathrm{N}}, \qquad
\mathbf{A}^{\mathrm{A}} = \frac{1}{k}\sum_{i=1}^{k}\mathbf{a}_i^{\mathrm{A}}.
\end{equation}

The semantic guidance loss pulls the learnable prompt features toward the corresponding anchor features:
\begin{equation}
\mathcal{L}_{\mathrm{sg}} = \frac{1}{2}\left(1-\cos(\mathbf{T}^{\mathrm{N}},\mathbf{A}^{\mathrm{N}}) + 1-\cos(\mathbf{T}^{\mathrm{A}},\mathbf{A}^{\mathrm{A}})\right).
\end{equation}

Diversity regularization penalizes pairwise similarity among prompt groups to encourage complementary semantics:
\begin{equation}
\mathcal{L}_{\mathrm{div}}^{\mathrm{type}} = \frac{1}{k(k-1)} \sum_{1 \le i < j \le k} \cos^2\left( \mathbf{t}_i^{\mathrm{type}}, \mathbf{t}_j^{\mathrm{type}} \right),
\end{equation}
with $\mathcal{L}_{\mathrm{div}} = \frac{1}{2}(\mathcal{L}_{\mathrm{div}}^{\mathrm{N}} + \mathcal{L}_{\mathrm{div}}^{\mathrm{A}})$. Orthogonality loss separates normal and abnormal embeddings:
\begin{equation}
\mathcal{L}_{\mathrm{orth}} = \left| \left\langle \mathbf{T}^{\mathrm{N}}, \mathbf{T}^{\mathrm{A}} \right\rangle \right|^2.
\end{equation}

\subsection{Training Objectives}
\label{ssec:loss}

Following the baseline~\cite{ma2025aaclip} for alignment (BCE for classification, Dice+Focal for segmentation) and orthogonality, the total loss additionally includes diversity and semantic guidance:
\begin{equation}
\mathcal{L}_{\mathrm{total}} = \lambda_{\mathrm{align}}\mathcal{L}_{\mathrm{align}} + \lambda_{\mathrm{div}}\mathcal{L}_{\mathrm{div}} + \lambda_{\mathrm{sg}}\mathcal{L}_{\mathrm{sg}} + \lambda_{\mathrm{orth}}\mathcal{L}_{\mathrm{orth}}.
\end{equation}

We set $\lambda_{\mathrm{align}}{=}1$, $\lambda_{\mathrm{div}}{=}0.15$, $\lambda_{\mathrm{sg}}{=}0.5$, $\lambda_{\mathrm{orth}}{=}0.1$.

\section{Experiments}
\label{sec:experiments}

\subsection{Setup}
We evaluate on 14 datasets: MVTec AD~\cite{bergmann2019mvtec}, VisA~\cite{zou2022spot}, BTAD~\cite{mishra2021vtadl}, MPDD~\cite{jezek2021deep}, DTD-Synthetic~\cite{aota2023zeroshot}, DAGM~\cite{wieler2007weakly}, and CVC-ClinicDB~\cite{bernal2015wmdova}, CVC-ColonDB~\cite{bernal2012towards}, Endo~\cite{hicks2021endotect}, TN3K~\cite{gong2021multitask}, HeadCT~\cite{salehi2021multiresolution}, BrainMRI~\cite{salehi2021multiresolution}, Br35H~\cite{tbkk-q937-25}, Kvasir~\cite{jha2019kvasir} (medical). Following the cross-dataset protocol, models are trained on VisA and evaluated on all others (for VisA, trained on MVTec AD). We use ViT-L-14-336 CLIP backbone and SAM2.1 hiera-large, with only adapters and prompts trainable. Metrics are pixel-level AUROC/AUPRO and image-level AUROC/AP.

\begin{table*}[htbp]
    \centering
    \caption{Image-level ZSAD (I-AUROC\% / AP\%)}
    \label{tab:image}
    \small
    \begin{tabular*}{\textwidth}{@{\extracolsep{\fill}}llcccccc@{}}
        \toprule
        Domain & Datasets & AnomalyCLIP & AdaCLIP & AA-CLIP & Bayes-PFL & MRAD & Ours \\
        \midrule
        & MVTec AD & (91.5,96.1) & (92.0,96.4) & (91.6,96.1) & (92.0,96.2) & (\textbf{94.0},\textbf{97.4}) & (\underline{93.8},\underline{97.1}) \\
        & VisA & (82.1,85.4) & (83.0,84.9) & (81.5,85.1) & (\textbf{87.0},\textbf{89.0}) & (\underline{85.7},\underline{88.3}) & (83.0,85.5) \\
        Industrial & BTAD & (88.3,87.3) & (91.6,92.4) & (\underline{93.2},\textbf{96.8}) & (92.8,94.1) & (92.8,94.2) & (\textbf{94.4},\underline{96.4}) \\
        & MPDD & (74.1,78.2) & (-,-) & (73.0,79.3) & (-,-) & (\textbf{81.8},\textbf{83.4}) & (\underline{77.8},\underline{82.5}) \\
        & DTD-Synthetic & (93.5,97.0) & (92.8,97.0) & (91.9,91.3) & (\textbf{97.2},\textbf{99.0}) & (\underline{96.0},\underline{98.4}) & (94.0,93.5) \\
        & DAGM & (97.5,92.3) & (96.5,\underline{95.7}) & (93.1,82.7) & (\underline{97.7},\underline{95.7}) & (\textbf{98.4},\textbf{98.6}) & (96.6,89.7) \\
        \midrule
        & BrainMRI & (90.3,92.2) & (\underline{94.4},93.1) & (91.5,91.3) & (94.3,88.4) & (\textbf{97.0},\textbf{97.4}) & (93.5,\underline{93.5}) \\
        Medical & HeadCT & (93.4,91.6) & (91.4,92.2) & (95.7,92.2) & (91.9,91.1) & (\textbf{97.1},\textbf{97.6}) & (\underline{96.7},\underline{96.5}) \\
        & Br35H & (94.6,94.7) & (96.1,94.3) & (95.3,94.1) & (\underline{97.8},93.6) & (\textbf{97.9},\textbf{97.6}) & (96.5,\underline{95.5}) \\
        \bottomrule
    \end{tabular*}
\end{table*}

% ===== 新增：Figure 2 及定性分析 =====
\begin{figure*}[t]
    \centering
    \includegraphics[width=\textwidth]{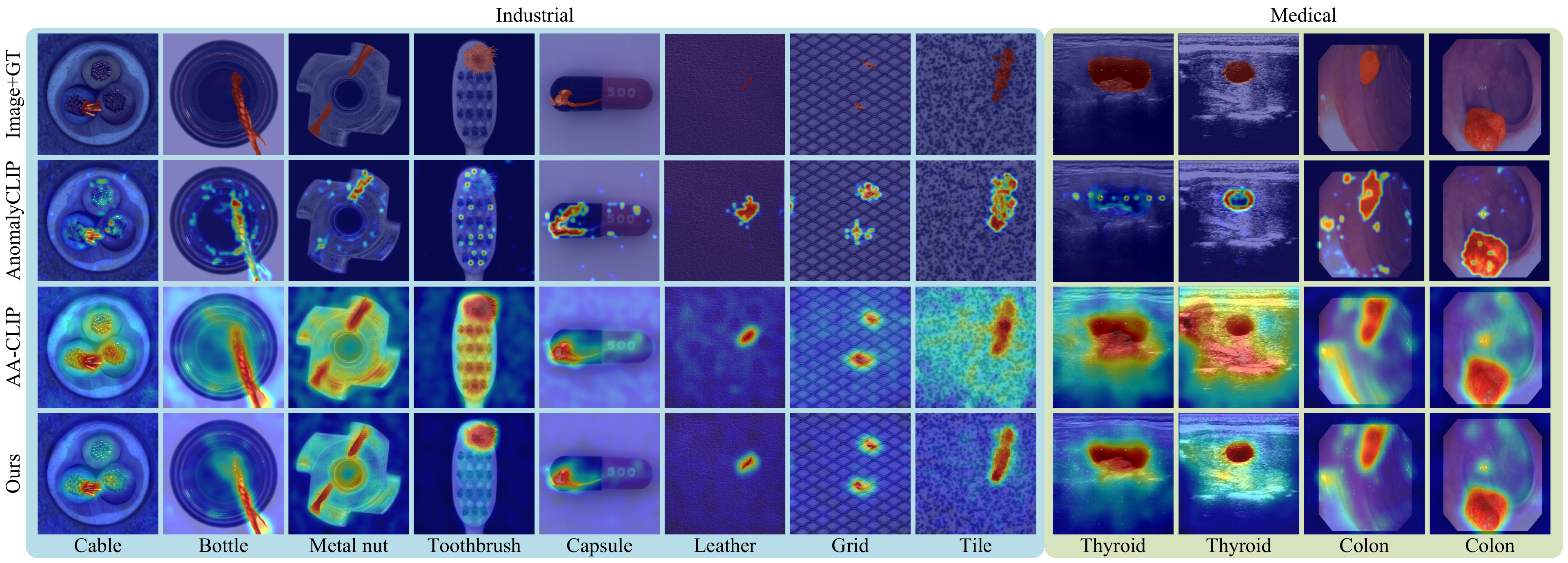}
    \caption{Qualitative comparison of anomaly segmentation across industrial (left) and medical (right) domains.}
    \label{fig:visual}
\end{figure*}

\subsection{Comparison with State-of-the-Art}
Tables~\ref{tab:pixel} and~\ref{tab:image} provide a full comparison with recent methods. Our method achieves the best pixel AUROC on MVTec AD (93.7), BTAD (95.5), DTD-Synthetic (98.6), CVC-ClinicDB (88.2), TN3K (84.8), Endo (91.3) and Kvasir (89.3), and best pixel AUPRO on CVC-ClinicDB (75.2), CVC-ColonDB (75.1), TN3K (51.9) and Endo (77.8). The improvements stem from PPSS providing precise boundary refinement and MSGPR enhancing coarse localization.

% ===== Table 3: 模块消融 (单栏，尝试就地放置) =====
\begin{table}[htbp]
    \centering
    \caption{Module ablation results. P-AUC/P-PRO averaged over 11 datasets with pixel-level GT and I-AUC/I-AP averaged over 9 datasets with image-level labels. Best results are highlighted in bold.}
    \label{tab:ablation}
    \small
    \begin{tabular}{lcccc}
        \toprule
        Method & P-AUC & P-PRO & I-AUC & I-AP \\
        \midrule
        Baseline (AA-CLIP) & 90.5 & 74.3 & 89.6 & 89.9 \\
        + PPSS & 91.3 & 76.5 & 91.7 & 92.0 \\
        + MSGPR & 90.9 & 75.0 & 90.4 & 90.7 \\
        + PPSS + MSGPR & \textbf{92.1} & \textbf{77.3} & \textbf{91.8} & \textbf{92.2} \\
        \bottomrule
    \end{tabular}
\end{table}

\subsection{Ablation Studies}
Table~\ref{tab:ablation} shows that adding PPSS alone improves P-AUC from 90.5 to 91.3 and P-PRO from 74.3 to 76.5, while I-AUC and I-AP rise to 91.7 and 92.0, indicating that SAM2 refinement helps both local and global discrimination. Adding MSGPR alone yields smaller pixel-level gains but improves image-level metrics to 90.4 and 90.7, mainly strengthening global semantics. Combining both achieves the best results (P-AUC 92.1, P-PRO 77.3, I-AUC 91.8, I-AP 92.2). The additional gains over individual components confirm the complementary synergy between PPSS and MSGPR.

Fig.~\ref{fig:visual} provides a visual comparison of anomaly segmentation results. In industrial scenarios, PSMP-CLIP yields clearer boundaries and better alignment with ground truth, even for subtle defects under complex backgrounds. Benefiting from rich multi-semantic prompts and patch-guided SAM2 refinement, the model identifies anomalies more accurately and suppresses false positives. These qualitative results further confirm the advantage of combining multi-semantic guided prompt regularization with fine-grained SAM2 segmentation.

\section{Conclusion}
\label{sec:conclusion}

We proposed PSMP-CLIP, a zero-shot anomaly detection network that synergizes patch-prompt SAM2 segmentation and multi-semantic prompt regularization. PPSS avoids prompt drift by sampling directly from patch anomaly scores, while MSGPR preserves general knowledge and enriches semantics. Experiments on 14 datasets show highly competitive pixel-level performance and strong cross-domain generalization. Future work will extend to video and few-shot settings.

% ===== 强制分页，使参考文献从新的一页开始 =====
\clearpage

\let\oldthebibliography\thebibliography
\let\endoldthebibliography\endthebibliography
\renewenvironment{thebibliography}[1]{%
  \oldthebibliography{#1}%
  \setlength{\itemsep}{3pt}%
  \setlength{\parskip}{3pt}%
  \setlength{\parsep}{3pt}%
}{\endoldthebibliography}

\bibliographystyle{IEEEbib}
\bibliography{refs}

\end{document}